\documentclass[runningheads]{llncs}
\usepackage{amsmath}
\usepackage[T1]{fontenc}
\usepackage{graphicx,verbatim}
\usepackage{booktabs}
\usepackage{amsfonts}
\usepackage{bm}
\usepackage{xcolor}
\usepackage{ulem}
\DeclareMathOperator*{\argmin}{arg\,min}
\def\xyz{\mathrm{x\hspace{-0.1em}y\hspace{-0.1em}z}}
\begin{document}
\title{DD-INR: Dynamics-Driven Implicit Neural Representation for Accelerated Whole-Brain Functional MRI Reconstruction}
\begin{comment}  %% Removed for anonymized MICCAI submission
\author{First Author\inst{1}\orcidID{0000-1111-2222-3333} \and
Second Author\inst{2,3}\orcidID{1111-2222-3333-4444} \and
Third Author\inst{3}\orcidID{2222--3333-4444-5555}}
%
\authorrunning{F. Author et al.}
\institute{Princeton University, Princeton NJ 08544, USA \and
Springer Heidelberg, Tiergartenstr. 17, 69121 Heidelberg, Germany
\email{lncs@springer.com}\\
\url{http://www.springer.com/gp/computer-science/lncs} \and
ABC Institute, Rupert-Karls-University Heidelberg, Heidelberg, Germany\\
\email{\{abc,lncs\}@uni-heidelberg.de}}

\end{comment}

\author{Qiaoxin Li\inst{1,2} \and
Caini Pan\inst{1,2,3} \and
Pierre-Antoine Comby\inst{1,2} \and
Chaithya Giliyar Radhakrishna\inst{1,2} \and
Philippe Ciuciu\inst{1,2}\thanks{Corresponding author}}
%index{Li, Qiaoxin}
%index{Pan, Caini}
%index{Comby, Pierre-Antoine}
%index{Radhakrishna, Chaithya Gilyar}
%index{Ciuciu, Philippe}
\authorrunning{Q. Li et al.}
\institute{MIND, Inria, Palaiseau, France \and
Neurospin, CEA Paris Saclay, France \and
CEA NeuroSpin, Paris-Saclay University, CNRS BAOBAB, Gif-sur-Yvette, France\\
\email{philippe.ciuciu@cea.fr}}
  
\maketitle              % typeset the header of the contribution
\begin{abstract}
Accelerated acquisition of fMRI enables enhanced detection of neurovascular (BOLD) activity in the brain, but image reconstruction becomes challenging with high k-space undersampling: Task-evoked BOLD signals are small in magnitude, which traditional anatomical MRI reconstruction methods fail to recover, as they favor spatial accuracy over temporal fidelity. We present DD-INR, a \emph{D}ynamics-\emph{D}riven \emph{I}mplicit \emph{N}eural \emph{R}epresentation framework tailored for accelerated fMRI that benefits from incoherent time-varying sampling and a tailored spatio-temporal prior, outperforming traditional methods, demonstrated in simulation and in-vivo acquisition, both in terms of image quality and retrieval of activation patterns.

DD-INR achieves this by splitting the fMRI data into a static background and a temporally varying dynamic component, representing only the dynamics with a dedicated INR, thereby focusing the model's capacity on activation-relevant changes while remaining compact.  
% Experiments across signal-to-noise ratios and acceleration rates, using both simulated and in vivo data, demonstrate consistent gains over strong reference methods and superior preservation of activation patterns. 
In general, DD-INR provides a promising framework for accelerated fMRI reconstruction, with the potential to improve the sensitivity and robustness of fMRI studies within practical scan time limits. 
\textit{The source code is available at \url{https://github.com/JoosenLi/DD-INR}.}
% Authors must provide keywords and are not allowed to remove this Keyword section.

\end{abstract}
\section{Introduction}

Functional Magnetic Resonance Imaging (fMRI) is pivotal for non-invasive brain mapping, yet it faces an inherent trade-off between spatial resolution, temporal resolution, and signal-to-noise ratio (SNR) \cite{faro2011functional}. 
High spatiotemporal resolution is critical to capture hemodynamic responses and improve the reliability of downstream tasks~\cite{aggarwal2024across,hu2024consecutive,huang2024topological}. However, acquiring high spatial and temporal resolution fMRI images requires strong k-space undersampling, making image reconstruction an ill-posed inverse problem. In particular, collecting high-speed fMRI data while maintaining full brain coverage requires time-varying k-space sampling patterns such as spiral stack or variations~\cite{petrov2017,comby2025snake}.

% missing prior work citations
% Check PAC thesis for "classical" references.
Solving the image reconstruction problem in fMRI requires introducing prior knowledge into reconstruction via traditional methods such as compressed sensing (CS) with spatial or temporal regularization \cite{amor2023noncartesian,chiew2018neuroimage,petrov2017}.
%Thus, to solve the problem, one must introduce prior knowledge into the reconstruction. Traditional methods include compressed sensing (CS) with regularization in the spatial and/or temporal domains \cite{chavarrias2015exploitation,chaari2014spatio}. 
More recently, deep learning~(DL) based methods have emerged, relying on supervised \cite{huang2021deep,sun2016deep} or self-supervised techniques \cite{korkmaz2023self,gu2024non,hu2021selfsupervised}, but their direct use in fMRI remains difficult: high-resolution ground-truth data for training are scarce, volumetric 3D+time data are computationally demanding, and hallucinated anatomical or functional features can be detrimental to downstream fMRI analysis. Solutions such as Plug-and-Play~(PnP) have shown promising results in fMRI~\cite{comby2025robust}, but remain limited to volume-wise processing.

%Other solutions include Plug-n-Play (PnP) methods\cite{comby2025robust} and diffusion-based models \cite{song2021solving,cao2024high,cui2024spirit} that have achieved state-of-the-art results in anatomical imaging and are now being tested in fMRI.

%exploit sparsity in transform domains but often suffer from hand-crafted priors and over-smoothing of fine-scale activations~\cite{chavarrias2015exploitation,chaari2014spatio}. Over the past decade, deep learning (DL) has emerged as a powerful alternative: Supervised unrolling networks and generative models, such as Plug-and-play(PnP) and diffusion models ~\cite{song2021solving,cao2024high,cui2024spirit}, have achieved state-of-the-art results in anatomical imaging. Despite their success, these methods face significant challenges in fMRI: they typically require large, fully-sampled training datasets—which are difficult to acquire for fMRI—and they risk hallucinating anatomical features that usually obscure subtle, task-evoked BOLD signals ~\cite{zhao2024tardrl}.
% TODO: mention that these methods don't scale for fMRI, and there is not supervised dataset. 
% TODO: Also due to the spatial-temporal resolution tradeoff, there is no ground truth with which one can compare and develop accelerated acquisition and reconstruction like in anatomical MRI.
% TODO: Also mention PnP in the introduction. 

To overcome data scarcity and generalization issues, Implicit Neural Representation~(INR) has gained traction \cite{zhu2025implicit,li2024acceleration,xu2025cardiac,shen2022nerp,shen2025imj}. INRs model images as continuous functions via coordinate-based networks, offering memory efficiency and optimization at reconstruction time from the measured data of a single subject rather than trained offline on a curated cohort. 
Although INRs have excelled in dynamic cardiac MRI~\cite{huang2023neural,chen2025single,feng2025spatiotemporal,baik2025dynamic,huang2025physics}, this application focuses mainly on resolving macroscopic anatomical deformations, targeting 2D+time sequences. However, fMRI must capture subtle and possibly fast fluctuations~\cite{lewis2016fast} in the intensity of the BOLD signal~(transient signal changes of only 2--5\% \cite{chang2008mapping}). This mismatch between reconstruction goals and the prohibitive memory requirements of volumetric representations renders the direct application of motion-centric INR approaches poorly suited for functional neuroimaging.

This paper introduces DD-INR, a \textit{Dynamics-Driven Implicit Neural Representation} framework explicitly tailored for accelerated fMRI reconstruction. 
Unlike generic dynamic INRs, DD-INR splits the 3D+time signal into a static anatomical background, reconstructed from temporally accumulated data, and a time-varying dynamic component, represented by a subject-specific INR. This decomposition allows the INR to focus its capacity on activation-relevant fluctuations rather than redundant anatomical structures that remain constant over time.
Furthermore, the differentiability of INRs is leveraged to enforce priors that encode spatial coherence of temporal changes, consistent with the hemodynamic responses observed in fMRI.

\section{Reconstructing fMRI data with INR}
\begin{figure}[hbt]
    \centering
    \includegraphics[width=\textwidth]{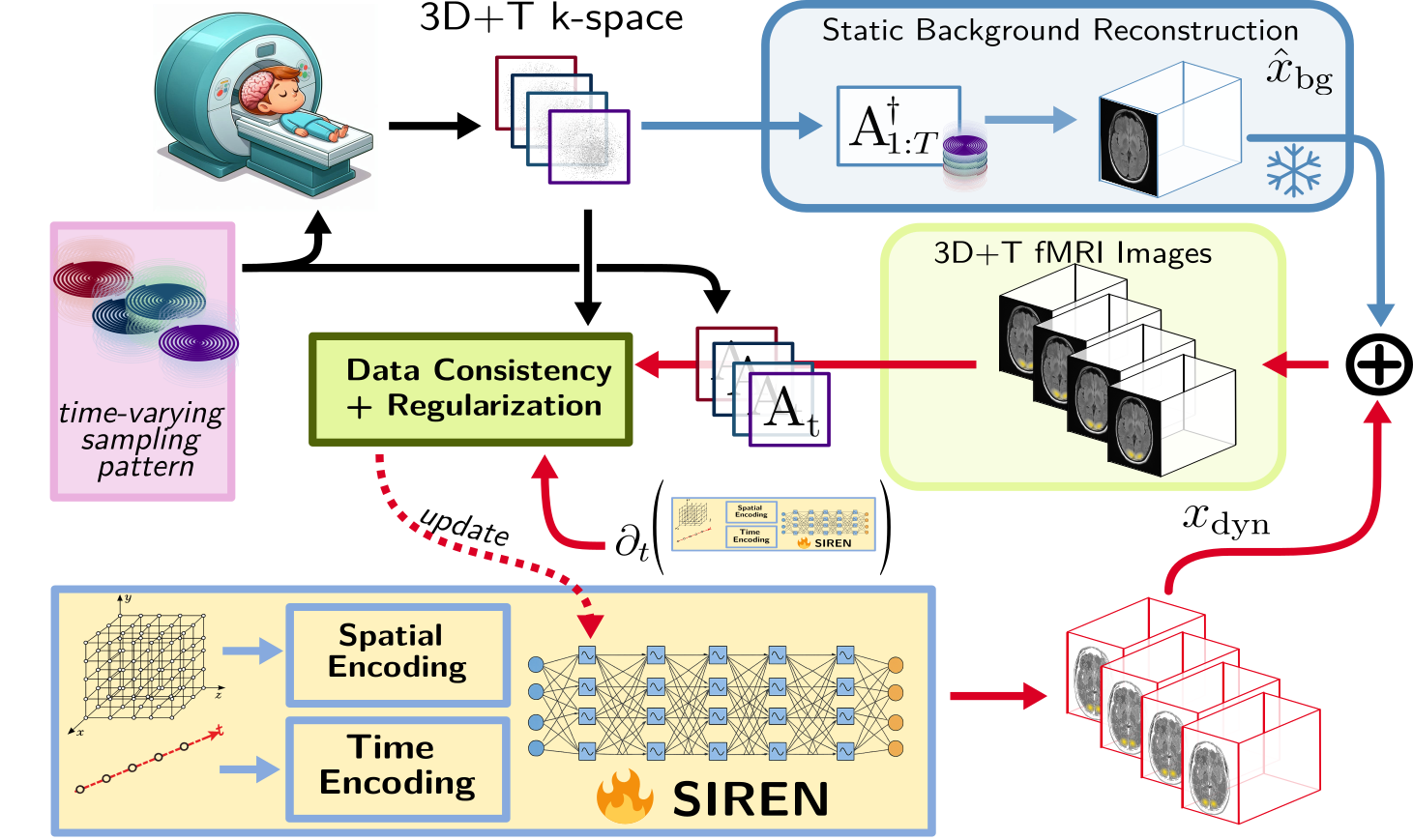}
    \caption{\label{fig:workflow}
    Architecture of proposed DD-INR for accelerated 3D+time~(3D+T) fMRI reconstruction with time-varying sampling. 3D+T fMRI data is acquired using a time-varying sampling pattern. For reconstruction, first a static background volume \(\hat{x}_{\rm bg}\) is reconstructed from samples pooled across time into a 3D k-space dataset. Then a SIREN INR is trained on the same subject to reconstruct the dynamic part \(x_{\rm dyn}\) encapsulating the BOLD signal. The training loop is highlighted with red arrows. The dynamic signal is regularized by spatial TV of the temporal derivative of the INR, computed via automatic differentiation of the SIREN block.
    }
\end{figure}
% \begin{figure}[htbp]
%     \includegraphics[width=\textwidth]{figures/workflow.jpg}
%     \caption{\label{fig:workflow}
% Architecture of proposed DD-INR for accelerated 3D+T fMRI reconstruction with time-varying sampling.
% \textbf{Left:} pipeline overview. Time-varying non-Cartesian k-space data are acquired (Step~1), a time-invariant background is reconstructed from temporally aggregated measurements (Step~2), and a dynamics-driven implicit neural representation reconstructs the time-varying component to form the final BOLD images (Step~3).
% \textbf{Middle:} acquisition and background reconstruction. Per-frame trajectories and k-space samples are stacked/accumulated across time to form an aggregated dataset, from which the static background is estimated using a CG solver.
% \textbf{Right:} dynamics reconstruction. Spatial and temporal coordinates are encoded by decoupled positional encoders and fed into a SIREN to represent the dynamic component as a continuous spatiotemporal field. The predicted dynamics are combined with the fixed background, mapped to k-space via NUFFT, and optimized using a data-consistency loss together with brain-informed regularization.}
% \end{figure}

\subsection{Reconstruction Formulation}

Reconstructing a sequence of fMRI volumes from accelerated multi-coil k-space data can be written as: 
\begin{equation} % with A_t
x = \argmin_{x \in \mathbb{C}^{N_\xyz \times T}}
\sum_{t=1}^{T}
\left\|A_{t} x_t-y_{t}\right\|_2^2
+ \lambda\mathcal{R}(x),
\end{equation}
\noindent where \(A_t\) represents the forward model~(e.g. non-uniform Fourier) operator in frame \(t\) (encapsulating the sampling pattern \(\Omega_t\) and the multi-coil encoding), for the associated \(y_t\) k-space data, and \(\lambda\mathcal{R}(x)\) is a regularization term promoting a given prior.

% \begin{equation} % with F and S
% x = \argmin_{x \in \mathbb{C}^{T\times N_\xyz}}
% \sum_{t=1}^{T}\sum_{\ell=1}^L
% \left\|\mathcal{F}_{\Omega_t}\mathcal{S}_\ell x_t-y_{t,\ell}\right\|_2^2
% + \mathcal{R}(x),
% \end{equation}
% where $x_t$ denotes the complex image at time $t$, $S_\ell$ the coil sensitivity map of the $\ell$-th coil, and $\mathcal{F}_{\Omega_t}$ the Fourier operator corresponding to the sampling trajectory $\Omega_t$. $\mathcal{R}(x)$ holds all regularization enforcing prior on the reconstruction. %Typical example are Wavelet, TV, Low Rank, Deep Image Prior... [CITES]
 
%\subsection{DD-INR: dynamics-driven implicit neural representation}

\subsubsection{Dynamics-driven signal factorization.}
In task-based fMRI, activation-related BOLD signal variations are small compared to the average intensity of the anatomical background. Directly modeling the full 3D+T signal, therefore, tends to spend most of the model capacity on time-invariant structure rather than on task-relevant dynamics. To better match the reconstruction objective, the signal at spatial coordinate \(r\) and time \(t\) is split into a background and a dynamic part:
\begin{equation}
x(r,t) = \hat{x}_{\mathrm{bg}}(r) + x_{\mathrm{dyn}}(r, t).
\end{equation}
The background is estimated from a temporally aggregated 3D k-space dataset. Specifically, the samples and trajectories \(\{y_t,\Omega_t\}_{t=1}^{T}\) are pooled across time, yielding \(y_{1:T}\) and a static forward operator \(A_{1:T}\):
\begin{equation}
\label{eq:bg-least-square}
\hat{x}_{\mathrm{bg}} =
\argmin_{z \in \mathbb{C}^{N_{\xyz}}}
\left\|A_{1:T}z-y_{1:T}\right\|_2^2 =  A_{1:T}^{\dagger}y_{1:T},
\end{equation}
where $A_{1:T}^\dagger$ denotes the Moore-Penrose inverse of the aggregated operator.

We now focus on parameterizing $x_{\mathrm{dyn}}$ using SIREN~\cite{sitzmann2020implicit}, accounting for spatial and temporal dimensions $f_\theta(r, t)$.
To improve the sensitivity to local variations in the fMRI signal of \(f_\theta\) , we use Fourier feature mapping embedding \cite{zheng2021rethinking}, separated in space and time:
\begin{align}
x_{\rm dyn}(r,t) &=  f_\theta(r,t) = \textsc{Siren}_{\theta}(\gamma_{\xyz}(r), \gamma_t(t)),\\
\gamma_{\xyz}(r)&=\left[\sin(2\pi B_\xyz r),\cos(2\pi B_\xyz  r)\right],\nonumber \\
\gamma_t(t)&=\left[\sin(2\pi B_t t),\cos(2\pi B_t t)\right].\nonumber
\end{align}
\noindent where \(B_\xyz\) and \(B_t\) are random Gaussian matrices. This decoupled design allows us to allocate different bandwidths and embedding dimensions, improving network capacity while reducing the overall parameter count and facilitating the recovery of subtle temporal dynamics.

\subsubsection{INR regularization.}
In task-based fMRI, the BOLD signal is expected to vary smoothly across neighboring voxels and to evolve coherently within activated regions. We therefore consider two Total-Variation (TV) penalties:
\begin{equation}
\mathcal{R}_{\xyz}(\theta) = 
\sum_{t=1}^{T}\mathrm{TV}_{\!\xyz}\!\left(x_{\mathrm{dyn}}(\cdot,t)\right)\quad \text{and}\quad
\mathcal{R}_{\xyz,\mathrm{t}}(\theta) = 
\sum_{t=1}^{T}\frac{1}{t}\,\mathrm{TV}_{\!\xyz}\!\left(\partial_t x_{\rm dyn}(\cdot,t)\right).
\end{equation}
\(\mathcal{R}_{\xyz}\) applies spatial TV directly to the dynamic residual at each time point. \(\mathcal{R}_{\xyz,\mathrm{t}}\) instead applies spatial TV to the temporal derivative, promoting spatial coherence in the temporal changes while preserving the anatomical structure as well as temporal changes. The intrinsic differentiable property of the INR makes \(\partial_t x_{\rm dyn}\) directly available by autodifferentiation, and the spatial components required for TV are computed using finite differences to reduce computational cost. 
% The regularization is motivated by two fMRI priors: activations are spatially clustered, and task-related hemodynamic changes evolve coherently across neighboring voxels. its temporal derivative
% $\partial_t x_{\mathrm{dyn}}(r,t)=\frac{\partial}{\partial t}f_\theta(\gamma_s(r),\gamma_t(t))$
% is directly available via automatic differentiation~\cite{paszke2017automatic}.
% To reduce computational cost in 3D reconstruction, spatial operators are implemented with discrete finite differences. The regularization term is instantiated as
% \begin{equation}
% \mathcal{R}_{\mathrm{dyn}}(\theta)=
% \begin{cases}
% \lambda_s \displaystyle\sum_{t=1}^{T}\mathrm{TV}_{\mathrm{spatial}}\!\left(x_{\mathrm{dyn}}(\cdot,t)\right), & \text{(spatial regularization)},\\[6pt]
% \lambda_{st} \displaystyle\sum_{t=1}^{T}\frac{1}{t}\,\mathrm{TV}_{\mathrm{spatial}}\!\left(\partial_t x_{\mathrm{dyn}}(\cdot,t)\right), & \text{(spatiotemporal regularization)},
% \end{cases}
% \end{equation}
% \subsection{Final Optimization}
% \noindent 
Therefore, dynamic reconstruction is obtained by optimizing
\begin{equation}
\mathcal{L}(\theta) = 
\sum_{t=1}^{T} \left\|A_t\left(\hat{x}_{\mathrm{bg}} + f_\theta(r,t)\right)-y_{t}
\right\|_2^2
+ \lambda \mathcal{R}(\theta),
\end{equation}
where $\mathcal{R}$ denotes $\mathcal{R}_{\xyz}$ or $\mathcal{R}_{\xyz, \mathrm{t}}$ depending on the chosen prior and \(\lambda>0\) is a regularization hyperparameter.

% TODO: Mention the training procedure, batching etc.

\section{Experimental Setup}
We evaluated the performance of DD-INR on both simulated and in vivo acquired data.
The simulated k-space data were generated using the SNAKE simulator \cite{comby2025snake}, which provides ground-truth hemodynamic activity localized on an anatomical phantom~\cite{aubert-broche_twenty_2006} consisting only of white matter, gray matter, and CSF to remain feasible on consumer hardware.
We simulated a 3T-like 3D gradient-echo sequence with parameters
($T\!R_{\mathrm{shot}} = 50 ms, T\!E = 30 ms,F\!A = 12^\circ$),
a field of view of $192\times192\times120$ mm$^3$, 3mm-isotropic resolution, \(T\!R_{\rm vol}\) = 1s, and total scan duration of 240 s, with a block-on/block-off (20s/20s) visual paradigm acquired with a stack-of-spiral trajectory (AF=2 in the \(k_z\) direction). The locations of the spirals in the stack were drawn randomly with a uniform distribution for each frame independently, while enforcing full sampling of the center 20\% of \(k_z\)~\cite{petrov2017}.

To evaluate reconstruction performance in a real-world setting, an in vivo experiment was conducted on an healthy volunteer using a 3T scanner (Siemens MAGNETOM Cima.X), following the same acquisition protocol as in simulation. The study was approved by the relevant local and national ethics committees, and written informed consent was obtained prior to participation. Dedicated 44-channel coil sensitivity maps were acquired before the fMRI run.  % CPP 100050
In simulation, SNR was defined in k-space following~\cite{comby2025snake} and tuned to match the multi-coil in vivo acquisition setting. 

\subsubsection{Implementation Details}
The forward model operators \(A_t\) were implemented using MRI-NUFFT\cite{comby2025mrinufft}.
% For the background reconstruction, the \((k,t)\)-space samples and trajectories were first pooled into a single 3D k-space dataset without a temporal dimension. 
The background component \(\hat{x}_{\rm bg}\) is first to be estimated, by aggregating all \((k,t)\)-space samples and trajectories, following Eq. \eqref{eq:bg-least-square} using  25 iterations of conjugate gradient (CG).

The dynamic component was modeled using a SIREN~\cite{sitzmann2020implicit} (3 layers of width 512) with spatial (\(B_{\xyz} \in \mathbb{R}^{256 \times 3}\)) and temporal encodings (\(B_t \in \mathbb{R}^{64}\)) and two output channels for the real and imaginary parts.  

% TODO: Add warm-up for learning rate.
Optimization was conducted with 1000 iterations of ADAM(learning rates $2\cdot10^{-5}$).  
Regularization weights were introduced using a warm-up schedule:  \(\lambda=0\) for the first 100 iterations, then increased linearly for the next 200 iterations (up to \(\lambda=10^{-2}\) for \(\mathcal{R}_{\xyz}\) and \(\lambda=2\cdot10^{-3}\) for \(\mathcal{R}_{\xyz,t}\)) 
% The typical reconstruction time for the 3\,mm simulation setting was approximately 2 hours on an NVIDIA RTX 6000 GPU, comparable with other traditional methods, and is dominated by the NUFFT computations~(i.e.\(A_t\)).

% Standard fMRI data analysis, including voxelwise regression and statistical tests, was performed using Nilearn~\cite{abraham2014machine}.  
% For in vivo data, motion parameters were estimated using SPM12. 

\subsubsection{Comparison Methods and Metrics}

DD-INR was compared against representative baselines spanning direct reconstruction, least squares reconstruction (CG), compressed sensing, model-based learning, and implicit neural representations. For fair comparison, competing methods used the   \(\hat{x}_{\rm bg}\) estimated by DD-INR as initialization or prior input whenever applicable. We compare to the following methods:
\textbf{NUFFT}: Adjoint NUFFT  without regularization, used as a reference baseline.
\textbf{CG}: \(\hat{x}_{bg}\)-initialized CG, representing an iterative baseline.
\textbf{CS}: \(\hat{x}_{bg}\)-initialized wavelet-regularized CS~\cite{amor2023noncartesian}, representing a sparsity-based baseline.
\textbf{PnP}: Plug-and-play reconstruction, specifically designed for fMRI~\cite{comby2025robust}, representing model-based learning approaches that incorporate learned priors.
\textbf{INR}: Prior-based INR~\cite{shen2022nerp,chen2025single}, used as a representative INR baseline; $\hat{x}_{bg}$ was provided for network initialization.

To analyze the activation detection accuracy,  GLM-based $z$-score maps were computed  and thresholded at $p<10^{-3}$. 
The resulting binary maps were compared with the simulated ground-truth activation region to compute F$_2$-scores, emphasizing recall and the detection of weak activation signals. Moreover, temporal fidelity was evaluated using the Pearson correlation coefficient (Corr) and normalized root mean square error (NRMSE) between the ROI-averaged reconstructed time series and the expected BOLD response. All metrics were computed over valid slices (defined as slices with more than 10 activated voxels in the ground-truth region) and reported as mean $\pm$ standard deviation. Corr reflects temporal consistency, while NRMSE quantifies amplitude accuracy.

% \subsubsection{Evaluation Metrics}

\section{Results}
\begin{figure}[bt]
    \centering
    \includegraphics[width=\textwidth]{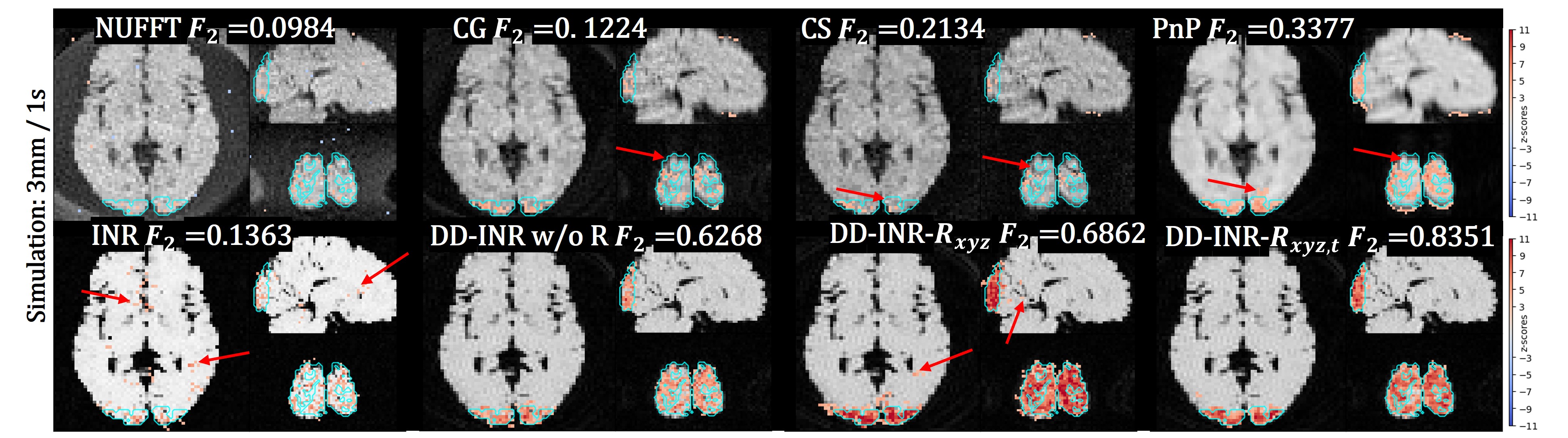}
    \caption{\label{fig:activation-map}
    GLM $z$-score activation maps from simulated 3D+T data. Columns compare NUFFT, CG, CS, PnP, INR, DD-INR without regularization, DD-INR with spatial regularization, and DD-INR with spatiotemporal regularization. Grayscale images show the magnitude of the first reconstructed frame (T$_2^*$), with three orthogonal views displayed. Cyan contours indicate the simulated ground-truth activation region in SNAKE. Red arrows highlight false-positive or false-negative activations, and blue arrows indicate artifacts in reconstructed T$_2^*$ images. The F$_2$-score is reported for each method.}
    \end{figure}
% Figure~\ref{fig:activation-map} compares activation maps and corresponding F$_2$-scores under two simulated acquisition settings. In the 3\,mm / 1\,s configuration, NUFFT, CG, and CS exhibit noisy reconstructions and limited overlap with the ground-truth activation region. PnP improves detectability but produces spatially diffuse activations, reflecting the trade-off between denoising strength and localization. The naive INR reconstructs anatomically plausible frames yet shows reduced activation sensitivity, suggesting insufficient fidelity of subtle temporal fluctuations when modeling the full 3D+T signal without dynamics-centric design.

% DD-INR improves alignment with the simulated activation region and achieves higher F$_2$-scores. Incorporating spatial regularization enhances stability but may slightly reduce localization sensitivity due to over-smoothing of dynamic variations. In contrast, the spatiotemporal regularization better preserves activation boundaries by constraining spatial coherence of temporal changes, resulting in improved balance between sensitivity and specificity.

% Under the more challenging 2\,mm / 1.5\,s setting, direct and sparsity-based reconstructions show degraded activation recovery. The INR baseline remains limited in capturing weak task-related fluctuations. DD-INR, particularly with spatiotemporal regularization, maintains artifact suppression while achieving the most accurate and spatially consistent activation maps, reflected in the highest F$_2$-scores across methods.
Fig.~\ref{fig:activation-map} compares the activation maps and the corresponding F2-scores in both simulation settings. NUFFT, CG, and CS exhibit noisy reconstructions with limited overlap with the reference activation map. PnP improves statistical sensitivity, but yields spatially diffuse patterns. The naive INR reconstructs anatomically plausible images but shows reduced statistical sensitivity, indicating insufficient recovery of subtle temporal dynamics when modeling the full 3D+T subject without dynamics-centric design.

DD-INR consistently improves alignment with the simulated activation region in the brain and achieves higher F$_2$-scores. Spatial regularization improves stability but slightly reduces localization precision, whereas spatiotemporal regularization better preserves activation boundaries. 
% sim2 : At higher spatial resolution (2\,mm / 1.5\,s), these differences become more pronounced, with DD-INR maintaining artifact suppression and achieving the most accurate and spatially consistent activation maps.

\begin{table}[bt]
\centering
\small
\setlength{\tabcolsep}{4pt}
\caption{\label{tab:roi_timeseries_metrics}
Quantitative comparison of ROI time-series recovery in simulation.
Correlation coefficient and NRMSE are computed against the expected BOLD response and reported as mean $\pm$ std over valid slices.
Best results are in \textbf{bold}, and second-best results are \underline{underlined}. 
DD-INR-$R_{\xyz,t}$ and DD-INR-$R_{\xyz}$ denote DD-INR with spatiotemporal and spatial regularization, respectively.}
{\scriptsize
\setlength{\tabcolsep}{2pt}
\begin{tabular}{lccccccc}
\toprule
 & DD-INR-$R_{\xyz,t}$ & DD-INR-$R_{\xyz}$ & INR & PnP & CS & CG & NUFFT \\
\midrule
Corr$\uparrow$ 
& \textbf{0.909}$\pm$0.058 
& \underline{0.903}$\pm$0.054 
& 0.40$\pm$0.16 
& 0.84$\pm$0.11 
& 0.55$\pm$0.13 
& 0.39$\pm$0.23 
& 0.39$\pm$0.16 \\
NRMSE$\downarrow$ 
& \textbf{0.136}$\pm$0.037 
& \underline{0.142}$\pm$0.033 
& 0.35$\pm$0.04 
& 0.17$\pm$0.05 
& 0.30$\pm$0.04 
& 0.35$\pm$0.07 
& 0.36$\pm$0.04 \\
\bottomrule
\end{tabular}}
\end{table}

\begin{figure}[t]
    \centering
    \includegraphics[width=\textwidth]{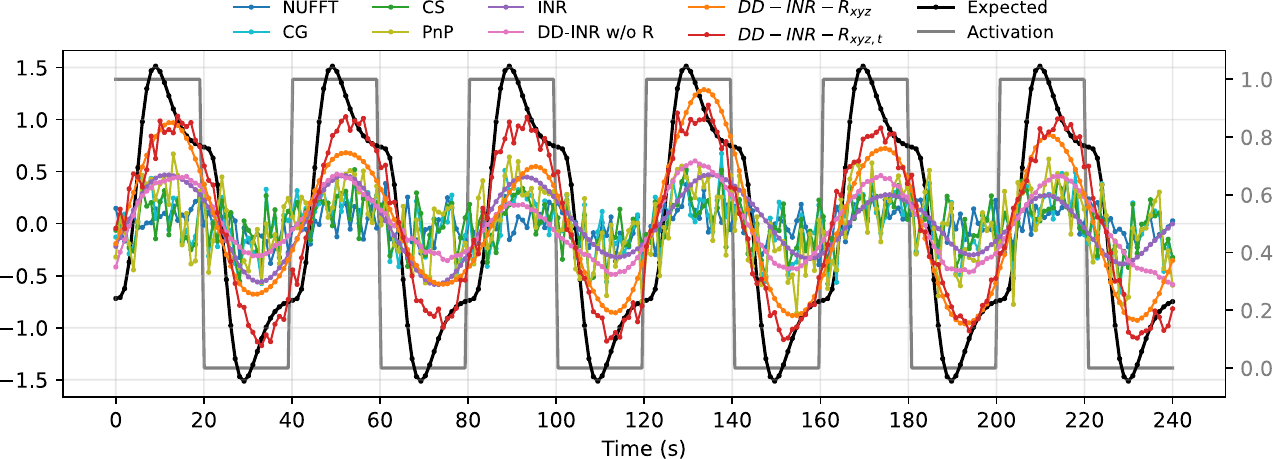}\\[-.35cm]
    \caption{\label{fig:time-series}
    ROI-averaged time series in the simulated activation region.
  %  Left: 3\,mm / 1\,s setting; Right: 2\,mm / 1.5\,s see expected BOLD response (gray) and block design (black) are shown for reference.
    Signals are normalized for visualization.}
\end{figure}

Fig.~\ref{fig:time-series} presents ROI-averaged time courses within the simulated activation region alongside the expected BOLD response. Most baselines capture coarse stimulus-locked trends but exhibit either pronounced temporal noise (NUFFT, CG, CS) or attenuated response amplitudes (PnP, naive INR). DD-INR provides a substantial improvement in temporal consistency and amplitude preservation.
Tab.~\ref{tab:roi_timeseries_metrics} further corroborates these observations quantitatively. DD-INR with spatiotemporal regularization achieves the highest correlation and lowest NRMSE.

% sim2: Under the more challenging setting 2\,mm/1.5\,s, temporal degradation becomes more evident for baseline methods, with stimulus-related oscillations partially distorted or suppressed. In contrast, DD-INR maintains coherent task-locked dynamics, highlighting the importance of explicit background–dynamics decomposition for recovering weak activation signals under stronger acceleration and reduced temporal resolution. 

\begin{figure}[tb]
    \centering
    \includegraphics[width=\textwidth]{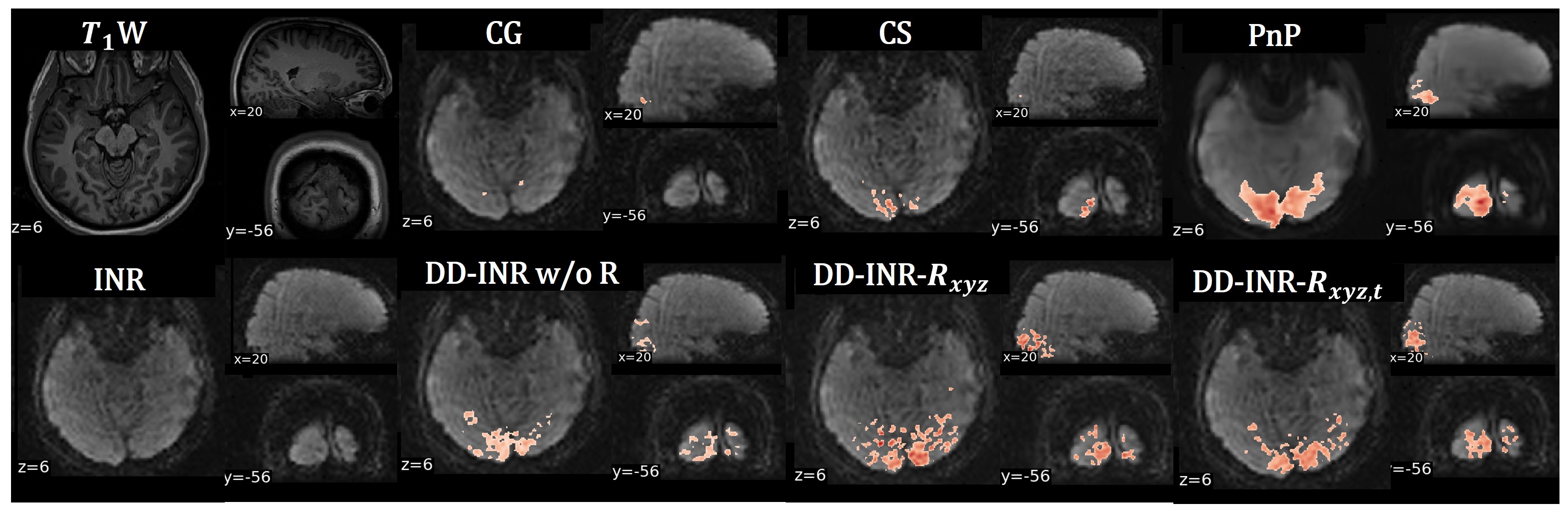}\\[-.25cm]
    \caption{\label{fig:real-activation}
    In vivo visual block-design results.2
    The top-left sub-figure shows the T1-weighted anatomical reference.
    Subsequent columns compare baselines and DD-INR with spatial / spatiotemporal regularization.
    Grayscale images correspond to reconstructed $T_2^*$ magnitude, with overlaid $z$-score activation maps. }
\end{figure}

Fig.~\ref{fig:real-activation} presents in vivo activation maps under the visual block-design paradigm. CG yields few activations, while CS provides limited improvement with small activation clusters. PnP produces broader activation patterns but exhibits over-smooth T$_2^*$ images, indicating reduced localization precision. The naive INR baseline reconstructs plausible anatomy, yet completely fails to detect task-locked activations.
In contrast, DD-INR shows consistent and spatially coherent activation patterns across views while preserving anatomical structure. % Incorporating brain-informed regularization further enhances activation detection. These findings suggest improved in vivo statistical sensitivity to evoked brain activity using DD-INR as an fMRI reconstruction method.

\section{Discussion and Conclusion}

\textbf{Simulation vs \textit{in vivo} experiment.}
The in vivo experiment yielded results consistent with the simulation, despite potentially being further degraded by uncontrollable factors such as physiological noise, scanner imperfections, and off-resonance artifacts. In the simulation, we quantitatively assessed the performance of DD-INR and provided a fair comparison with state-of-the-art fMRI reconstruction methods. Both in vivo and simulation studies show that spatial or spatiotemporal regularization improves the detection and strength of brain activation after standard analysis.

\textbf{Downstream fMRI inference.}
Spatiotemporal regularisation may alter temporal correlations, residual variance , and effective degrees of freedom, which in turn can skew the GLM analysis. To control for this effect, our simulation evaluates GLM-derived activation maps against a known ground truth, in addition to ROI time-series fidelity. Under this setting, DD-INR and its regularizers increase statistical sensitivity while preserving reasonable spatial specificity. Nevertheless, further work is needed to quantify reconstruction-induced activation spread and to understand how such effects impact downstream analysis.

\textbf{Spatio-temporal vs Spatial regularization.}
Both spatial and spatio-temporal regularization improve BOLD recovery in the INR reconstruction. Spatial-only regularization promotes image-domain smoothness, which can blur activation boundaries, spread task-related dynamics, or partially suppress weak BOLD components.  Instead, the spatio-temporal regularization encourages the spatial coherence of temporal variations, allowing neighboring voxels to exhibit consistent dynamic changes while preserving local spatial texture and temporal response amplitude. As a result, it reduces incoherent fluctuations and mitigates the diffuse activation patterns observed with spatial-only regularization.

% Both regularization on the INR improve BOLD recovery. However, the spatial-only prior introduces slightly diffuse activation patterns. The spatio-temporal regularization mitigates this effect, at the cost of an extra temporal derivative computation.

%but with different trade-offs. The spatial prior is faster and computationally simpler, but introduces slightly diffuse activation patterns due to the smoothing of the dynamic component. In contrast, the spatiotemporal prior regularizes the spatial variation of the temporal derivative, but it requires an additional temporal derivative through the INR. It is also more closely aligned with the hemodynamics of fMRI signals.

\textbf{Impact of movement.}
DD-INR assumes a time-invariant background; therefore, any potential residual head motion has to be accommodated by the INR component, and head motion can violate this assumption. The consistent in vivo activation patterns observed in this work suggest that the proposed factorization is practical under routine fMRI motion levels. 

\textbf{Relation to previous INR reconstructions.}
To our knowledge, DD-INR is the first INR framework explicitly tailored for accelerated fMRI. %, where the reconstruction objective is dominated by detecting weak BOLD fluctuations rather than anatomical structure or motion.
Most existing INR-based dynamic MRI reconstructions model the full spatiotemporal signal (or a latent subspace that captures both anatomy and dynamics) and are primarily developed for motion-dominated applications~\cite{huang2023neural,feng2025spatiotemporal,chen2025single,baik2025dynamic}, often with periodic dynamics (e.g., cardiac CINE). Chen et al.~\cite{chen2025single} leverage temporally aggregated measurements to stabilize training, but the INR is still required to model both static anatomy and temporal variations.
In contrast, in fMRI, the goal is to recover weak BOLD fluctuations superimposed on a strong static background, rather than to estimate the fine anatomical structure or motion. DD-INR is therefore explicitly problem-driven: It reconstructs a static background from temporally aggregated data and uses the INR capacity to represent only the dynamic residual, which better matches the objective of activation recovery. 

% Finally, while decomposition-based formulations (e.g., low-rank plus sparse) are effective for periodic motion dynamics in cardiac imaging, their underlying assumptions are less aligned with fMRI, where task-evoked BOLD responses are not periodic and do not primarily correspond to motion drifts. DD-INR therefore adopts a problem-driven background--dynamics factorization and hemodynamic-informed regularization designed specifically for preserving activation-related dynamics in accelerated fMRI.
\textbf{Conclusions.}
This work introduces DD-INR, a dynamic-driven implicit neural representation framework for accelerated fMRI that adapts signal representation, network architecture, and regularization to the dynamics of fMRI, concentrating model capacity on activation-related fluctuations. Experiments on simulated and in vivo data demonstrate improved activation detection and temporal fidelity under acceleration, highlighting the value of dynamics-centric modeling for more sensitive and robust accelerated fMRI. 

\begin{credits}
\subsubsection{\ackname}
This work is funded by WP2 (Optimizing Workflows Efficiency) of The European Joint Virtual Lab on Artificial Intelligence, Data Analytics and Scalable Simulation (AIDAS), an initiative of Forschungszentrum Jülich (FZJ) and the French Alternative Energies and Atomic Energy Commission (CEA).
\subsubsection{\discintname}
The authors have no competing interests to declare that are relevant to the content of this article.
\end{credits}


\bibliographystyle{splncs04}
\bibliography{mybibliography}
%
% IT would be best to have DOI as well, at least for us so that we can check references.

\end{document}